\documentclass{article}

\usepackage{PRIMEarxiv}

\usepackage[utf8]{inputenc} 
\usepackage[T1]{fontenc}    
\usepackage{hyperref}       
\usepackage{url}            
\usepackage{booktabs}       
\usepackage{amsfonts}       
\usepackage{nicefrac}       
\usepackage{microtype}      
\usepackage{lipsum}
\usepackage{fancyhdr}       
\usepackage{graphicx}       
\usepackage{amsmath}
\usepackage{float}
\graphicspath{{media/}}     
\usepackage{enumitem}

\title{AgroSense 2.0: Cross-Modal Transformer Fusion with Geospatial Raster Integration and Interpretable Multi-Task Learning for Precision Crop Recommendation}

\author{
  Vishal Pandey\\
  Research Engineer\\
  London, UK\\
  \texttt{vishal@metriqual.com} \\
  \And
  Rishav Tewari \\
  Independent Researcher\\
  Kolkata, IN\\
  \texttt{rishavtewari.research@gmail.com} \\
  \AND
  Ruzina Haque Laskar\\
  Scientist-B\\
  Centre for Development of Telematics\\
  Delhi, IN\\
  \texttt{ruzinah@cdot.in} \\
}

\begin{document}
\maketitle

\begin{abstract}
Crop recommendation systems in precision agriculture have long suffered from a fundamental modality gap: visual soil characterization and chemical nutrient profiling are typically treated as independent inference problems, with fusion often reduced to late-stage feature concatenation. AgroSense~2.0 addresses this limitation through three key architectural advances over its predecessor. First, we introduce continental-scale geospatial integration by incorporating a seven-band soil raster (\texttt{india\_soil\_7bands.tif}) spanning the entirety of India. The raster encodes Nitrogen, pH, Soil Organic Carbon (SOC), Clay, Sand, Silt, and Bulk Density, which are extracted as spatial patches of size $32 \times 32$ and used as an additional modality. This geospatial information was entirely absent in prior work and provides localized environmental context for downstream prediction tasks. Second, we replace naive feature concatenation with a cross-modal Transformer fusion module. In this architecture, tabular nutrient features attend over image representations using multi-head attention, enabling the model to learn complex inter-modal dependencies at the feature level rather than relying on shallow fusion strategies. This design facilitates richer interactions between chemical soil properties and visual-geospatial characteristics. Third, we adopt a multi-task learning objective that jointly optimizes soil classification and crop recommendation through a shared backbone network. By leveraging shared representations across related tasks, the model improves generalization and captures complementary information useful for both objectives. To enhance interpretability, we perform TreeSHAP analysis on the tabular branch of the framework. The resulting explanations reveal crop-conditioned nutrient sensitivity patterns. For example, rainfall and pH emerge as the most influential features globally, whereas potassium and temperature are identified as the primary drivers for coffee cultivation recommendations. These insights provide transparency into the model's decision-making process and align with established agronomic knowledge. Together, these contributions establish AgroSense~2.0 as a more principled, interpretable, and geospatially grounded framework for real-world precision agriculture, bridging the gap between visual soil understanding and nutrient-aware crop recommendation.
\end{abstract}


\section{Introduction}

Precision agriculture demands inference systems capable of reasoning simultaneously 
over the physical structure of soil, its chemical composition, and its spatial 
distribution across a landscape. Despite rapid advances in deep learning for 
agricultural decision support~\cite{garcia2024, lakshmi2025}, the dominant paradigm in crop recommendation continues to treat these three information sources as independent: visual soil characteristics are classified in isolation, nutrient profiles are modeled from point-sample tabular records, and fusion, when attempted at all, is performed through late-stage feature concatenation~\cite{dey2024, 
kaur2025}. This architectural choice is not merely suboptimal; it is fundamentally 
at odds with the structure of the problem. Soil properties are spatially continuous 
fields, not independent scalars. The nitrogen content of a pixel is correlated with its neighbors, its clay fraction, and its organic carbon, dependencies that a 
concatenation-based fusion module, operating on flattened feature vectors, cannot 
capture.

Our earlier work, AgroSense~\cite{agrosense_v1}, demonstrated the viability of multimodal fusion for crop recommendation, achieving 98.0\% accuracy by combining ResNet-18, EfficientNet-B0, and Vision Transformer classifiers with LightGBM-based nutrient profiling. However, AgroSense suffered from three structural limitations. 
First, its fusion mechanism was a simple feature concatenation, the one-hot encoded 
soil type was appended to the nutrient vector and passed to a downstream classifier, with no mechanism to model cross-modal interactions. Second, its tabular nutrient data consisted of point samples drawn from curated CSV repositories, discarding the spatial continuity that makes soil a fundamentally geospatial phenomenon. Third, while the system produced accurate recommendations, it offered no interpretability into \emph{which} soil or climate features drove predictions for \emph{which} crops, a critical requirement for farmer-facing deployment, where trust is predicated on 
explanation.

AgroSense 2.0 addresses each of these limitations through targeted architectural 
and data-level interventions, making the following contributions:

\begin{itemize}
    \item \textbf{Geospatial Raster Integration.} We introduce a 7-band continental-scale 
    soil raster (\texttt{india\_soil\_7bands.tif}) covering Nitrogen, pH, Soil Organic 
    Carbon (SOC), Clay, Sand, Silt, and Bulk Density across India. Spatial patches of 
    size $32 \times 32$ pixels are extracted with stride 16 and per-channel z-score 
    normalization, transforming the tabular point-sample paradigm into a spatially 
    continuous representation. To our knowledge, this is the first crop recommendation system to incorporate continental-scale multi-band soil raster data as a primary 
    input modality.

    \item \textbf{Cross-Modal Transformer Fusion.} We replace late-stage concatenation with a cross-modal attention module in which tabular nutrient features serve as \emph{queries} attending over image-derived \emph{keys} and \emph{values}. This allows the model to selectively weight visual soil features based on the chemical context of each sample, a directional, asymmetric interaction that concatenation cannot express. The fusion module uses 8-head attention with a shared 256-dimensional projection space for both modalities.

    \item \textbf{Multi-Task Learning.} AgroSense 2.0 jointly optimizes soil type classification and crop recommendation under a single shared EfficientNet-B0 backbone, using a weighted multi-task loss $\mathcal{L} = \mathcal{L}_{\text{crop}} + \lambda \mathcal{L}_{\text{soil}}$ with $\lambda = 0.3$. This regularizes the visual encoder toward semantically meaningful soil representations, improving generalization of the crop recommendation head.

    \item \textbf{Interpretability via TreeSHAP.} We apply TreeSHAP~\cite{lundberg2020} 
    to the LightGBM tabular branch, producing global feature importance rankings, per-crop attribution profiles, and a full $22 \times 7$ crop-feature SHAP heatmap. This analysis reveals agronomically meaningful patterns, rainfall and pH emerge as globally dominant features, while potassium and temperature are primary drivers for coffee, and humidity and nitrogen govern rice recommendations, providing the explanatory grounding necessary for real-world deployment.
\end{itemize}

Together, these advances position AgroSense 2.0 as a principled step toward  geospatially-grounded, interpretable, and architecturally rigorous multimodal learning for precision agriculture. The remainder of this paper is organized as follows: Section~\ref{sec:related} reviews related work; Section~\ref{sec:method} describes the full methodology; Section~\ref{sec:results} presents experimental results and ablation studies; Section~\ref{sec:discussion} discusses limitations and future directions.

\section{Related Work}
\label{sec:related}

\subsection{Multimodal Fusion in Precision Agriculture}

Early crop recommendation systems operated exclusively on structured tabular 
inputs, soil nutrient levels, pH, and climate variables, using classical 
machine learning methods such as Random Forests, SVMs, and gradient-boosted 
trees~\cite{motwani2022, rajak2017}. While effective on clean, laboratory-sourced 
data, these approaches are fundamentally limited by their inability to incorporate 
the visual and spatial characteristics of soil that agronomists routinely use in 
field diagnosis.

The introduction of convolutional neural networks into soil analysis opened a  parallel track of image-based classification~\cite{manjula2022}, but the two streams visual and chemical remained largely disconnected. Dey and 
Sharma~\cite{dey2024} represent the closest prior work to AgroSense 2.0 in 
spirit: their Agro-Deep Learning Framework fuses CNN-extracted visual features 
with gradient-boosted nutrient predictors, reporting 85.4\% accuracy and an 
88.9\% F1-score. However, their fusion mechanism is, again, late-stage 
concatenation, the CNN and the tabular model operate independently, with outputs 
merged only at the decision boundary. Shamsuddin et al.~\cite{shamsuddin2024} 
take a notably different approach, integrating UAV LiDAR point clouds, 
hyperspectral time-series, and weather data through an attention-based architecture 
for early maize yield prediction. Their work establishes that attention mechanisms 
can meaningfully arbitrate between heterogeneous agricultural modalities, a 
finding that motivates our cross-modal design, but their input modalities and 
task formulation differ substantially from the soil-to-crop recommendation setting. 
Liu et al.~\cite{liu2024} similarly demonstrate attention-based fusion of 
hyperspectral and LiDAR data for crop performance forecasting, reinforcing the 
pattern that inter-modal attention consistently outperforms feature concatenation in agricultural multimodal settings.

Critically, none of these works incorporate continental-scale geospatial soil raster data as a primary input modality. The shift from point-sample tabular records to spatially continuous multi-band rasters represents a fundamentally different data regime, one that captures the spatial autocorrelation of soil properties that point samples, by construction, cannot.

\subsection{Cross-Attention for Vision-Tabular Fusion}

The broader machine learning literature has increasingly recognized that 
transformer-based attention mechanisms offer a principled solution to the vision-tabular fusion problem. Gorishniy et al.~\cite{gorishniy2021} introduced 
the Feature Tokenizer Transformer (FT-Transformer), which applies self-attention 
across tokenized tabular features, substantially outperforming tree-based methods on structured benchmarks. Their work established that transformers are not merely 
a vision or language tool, they can model complex feature interactions in tabular 
data with comparable or superior effectiveness. AgroSense 2.0 extends this 
intuition into the cross-modal setting: rather than self-attention over tabular 
tokens, we employ \emph{cross}-attention in which tabular nutrient features query 
image-derived representations, enabling the chemical context of a soil sample to 
selectively gate which visual features are attended to. This asymmetric, 
directional interaction, tabular queries over visual keys and values, is 
architecturally distinct from both FT-Transformer's unimodal self-attention and 
the symmetric fusion used in prior agricultural systems.

Recent vision-language models~\cite{radford2021} have demonstrated the expressiveness of cross-modal attention at scale; our contribution is to adapt 
this principle to a domain where one modality is structured numerical data rather 
than natural language, and where the downstream task is agronomic rather than 
semantic retrieval.

\subsection{Interpretability in Agricultural AI}

The deployment gap between laboratory-accurate crop recommendation models and 
farmer-facing tools is, in large part, an interpretability gap. Turgut et 
al.~\cite{turgut2024} address this directly in AgroXAI, an explainable-AI crop 
recommender that integrates LIME and SHAP for transparent decision support, 
demonstrating that explanation quality is as important as predictive accuracy for 
stakeholder adoption. Li et al.~\cite{li2023} provide a comprehensive review of label-efficient learning in agriculture, noting that interpretability tools substantially reduce the annotation burden by enabling domain experts to identify and correct systematic model errors. AgroSense 2.0 adopts TreeSHAP~\cite{lundberg2020} 
specifically for its exactness, unlike LIME's local approximations, TreeSHAP 
computes exact Shapley values for tree-based models in polynomial time, producing 
globally consistent feature attributions. Applied to our LightGBM tabular branch, 
this yields not only global feature rankings but crop-conditioned attribution 
profiles across all 22 crop classes, a granularity of interpretability that, to 
our knowledge, has not been reported in the crop recommendation literature.

\noindent\textbf{Positioning:} AgroSense 2.0 sits at the intersection of these 
three research threads: it advances multimodal agricultural fusion through 
geospatial raster integration, replaces concatenation with principled cross-modal 
attention grounded in the FT-Transformer lineage, and delivers crop-level 
interpretability through exact Shapley attribution. No prior work unifies all 
three.

\section{Methodology}
\label{sec:method}

\subsection{Dataset and Geospatial Integration}

AgroSense 2.0 operates over three distinct input modalities, each contributing 
a complementary view of soil state. We describe each in turn, with explicit 
attention to what each modality adds over the point-sample tabular paradigm 
of AgroSense v1~\cite{agrosense_v1}.

\paragraph{Modality I: Continental-Scale Geospatial Soil Raster:} The primary data innovation of AgroSense 2.0 is the incorporation of 
\texttt{india\_soil\_7bands.tif}, a multi-band GeoTIFF raster covering the 
Indian subcontinent at spatial resolution sufficient for regional agronomic 
analysis. The raster encodes seven soil property channels: Nitrogen (N), 
pH, Soil Organic Carbon (SOC), Clay fraction, Sand fraction, Silt fraction, 
and Bulk Density , each as a spatially continuous field over the raster 
grid. This is a fundamentally different data structure from the point-sample 
CSV records used in AgroSense v1. Where tabular nutrient samples treat soil 
properties as independent scalar observations drawn from discrete field 
locations, the raster encodes the spatial autocorrelation of soil properties 
, the fact that nitrogen content at a given pixel is statistically dependent 
on its neighbors, its clay fraction, and its organic carbon concentration. 
This spatial continuity is agronomically real and statistically exploitable; 
ignoring it, as point-sample systems do, discards information that is both 
freely available and physically meaningful.

Patch extraction proceeds as follows. Let $\mathbf{R} \in \mathbb{R}^{C 
\times H \times W}$ denote the full raster array, where $C = 7$ bands and 
$(H, W)$ are the spatial dimensions. We extract non-overlapping-biased patches 
of size $P = 32$ with stride $S = 16$:

\begin{equation}
    \mathbf{p}_{y,x} = \mathbf{R}_{:,\; y:y+P,\; x:x+P}, 
    \quad y \in \{0, S, 2S, \ldots\},\; x \in \{0, S, 2S, \ldots\}
\end{equation}

Patches where $\mathbb{E}[\mathbf{p}_{y,x}] \leq 0$ are discarded as 
predominantly empty (ocean, masked, or no-data regions). We retain a maximum 
of $N_{\text{patch}} = 5{,}000$ valid patches, yielding a dataset of shape 
$(5000, 7, 32, 32)$. Each channel $c$ is independently normalized via 
per-channel z-score normalization:

\begin{equation}
    \hat{\mathbf{p}}^{(c)} = \frac{\mathbf{p}^{(c)} - \mu_c}{\sigma_c + 
    \epsilon}, \quad \mu_c = \mathbb{E}[\mathbf{p}^{(c)}],\; 
    \sigma_c = \text{Std}[\mathbf{p}^{(c)}],\; \epsilon = 10^{-6}
\end{equation}

This per-channel normalization is essential: the seven bands occupy 
heterogeneous physical scales (e.g., Bulk Density in g/cm$^3$ vs.\ pH on a 
logarithmic scale), and joint normalization would distort the relative 
magnitudes of agronomically meaningful signals.

\paragraph{Modality II: CycleGAN-Augmented Soil Image Dataset:} We use a 7-class soil image dataset comprising the following categories: Alluvial, Arid, Black, Laterite, Mountain, Red, and Yellow soils. The dataset is provided in two variants: an \texttt{Orignal-Dataset} of photographically sourced images, and a \texttt{CyAUG-Dataset} generated via CycleGAN-based unpaired image-to-image translation~\cite{zhu2017}. 
The CyAUG variant substantially expands per-class sample counts by 
synthesizing visually plausible soil images through learned style transfer 
across soil type domains. We train and evaluate AgroSense 2.0 on the CyAUG 
variant, as its larger scale and intra-class diversity better support 
learning from scratch , a necessary constraint given the unavailability 
of pretrained ImageNet weights in the Kaggle execution environment. All 
images are resized to $224 \times 224$ pixels. Standard photometric 
augmentations are applied during training: random horizontal flipping, 
random rotation up to $20^\circ$, and color jitter with brightness 
perturbation of $\pm 0.2$.

\paragraph{Modality III: Nutrient Tabular Data:} The tabular modality consists of structured soil and climate records with 
seven input features: Nitrogen (N), Phosphorus (P), Potassium (K), 
Temperature, Humidity, pH, and Rainfall. The target variable encodes 
22 crop classes, label-encoded via \texttt{LabelEncoder} and one-hot 
decoded at inference. Features are standardized using a \texttt{StandardScaler} 
fitted on the training split. The dataset is partitioned using stratified 
sampling at an 80/10/10 train/validation/test ratio to preserve class 
balance across all splits.

\paragraph{Cross-Modal Pairing Strategy:} Since the three modalities are sourced from distinct repositories without 
shared geographic identifiers, true geo-registration is not available. 
We adopt a stochastic cross-modal sampling strategy during training: for 
each image index $i$, the corresponding tabular sample is drawn as 
$\text{tab\_idx} = i \bmod |\mathcal{D}_{\text{tab}}|$, cycling through 
the nutrient dataset. This ensures balanced modality exposure across 
training batches. We acknowledge this as a limitation and discuss 
geo-registered pairing as a priority for future work 
(Section~\ref{sec:discussion}).


\subsection{AgroSense 2.0 Architecture}

\begin{figure}[H]
    \centering
    \includegraphics[width=0.9\linewidth]{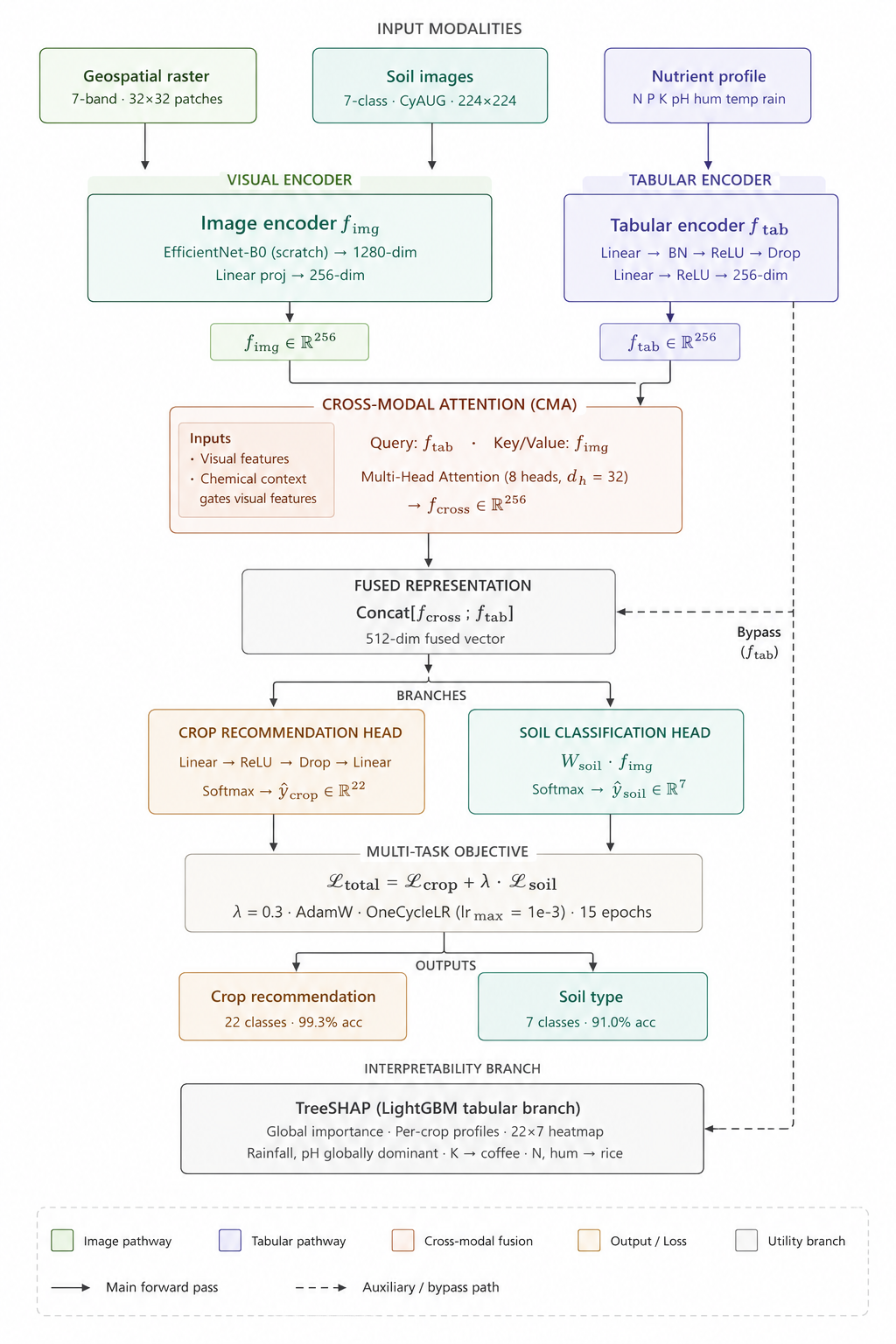}
    \caption{
        \textbf{AgroSense 2.0 Architecture.} Soil images and tabular nutrient 
        vectors are processed through independent encoders projecting to a shared 
        256-dimensional space. Tabular features attend over image features via 
        8-head cross-modal attention. The fused representation drives the crop 
        recommendation head, while a parallel soil classification head is 
        supervised directly from image features under the multi-task objective.
    }
    \label{fig:architecture}
\end{figure}

The central architectural contribution of AgroSense 2.0 is the replacement 
of late-stage feature concatenation with a cross-modal transformer fusion 
module. The full architecture, illustrated in Figure~\ref{fig:architecture}, 
comprises four components: an image encoder, a tabular encoder, a cross-modal 
attention module, and dual task-specific output heads.

\paragraph{Image Encoder $f_{\text{img}}$:} We use EfficientNet-B0~\cite{tan2019} as the visual backbone, initialized 
from random weights (no pretrained initialization). The classifier head is 
removed, exposing the 1280-dimensional penultimate feature vector. A learned 
linear projection $\mathbf{W}_{\text{img}} \in \mathbb{R}^{1280 \times d}$ 
maps this to the shared embedding space of dimension $d = 256$:

\begin{equation}
    \mathbf{f}_{\text{img}} = \mathbf{W}_{\text{img}}\; 
    \phi_{\text{EffNet}}(\mathbf{x}_{\text{img}}) \in \mathbb{R}^{d}
\end{equation}

where $\phi_{\text{EffNet}} : \mathbb{R}^{3 \times 224 \times 224} \to 
\mathbb{R}^{1280}$ denotes the backbone feature extractor. Training 
EfficientNet-B0 from scratch, while unconventional, is necessitated by 
the domain specificity of soil imagery , natural image priors from ImageNet 
may not transfer beneficially to the spectral and textural characteristics 
of soil photography.

\paragraph{Tabular Encoder $f_{\text{tab}}$:} The tabular branch maps the 7-dimensional normalized nutrient vector 
$\mathbf{x}_{\text{tab}} \in \mathbb{R}^{7}$ to the shared embedding 
space through a feed-forward network with batch normalization:

\begin{equation}
    \mathbf{f}_{\text{tab}} = \text{ReLU}\!\left(\mathbf{W}_2\; 
    \text{ReLU}\!\left(\text{BN}\!\left(\mathbf{W}_1\, 
    \mathbf{x}_{\text{tab}}\right)\right)\right) \in \mathbb{R}^{d}
\end{equation}

where $\mathbf{W}_1 \in \mathbb{R}^{128 \times 7}$, $\mathbf{W}_2 \in 
\mathbb{R}^{d \times 128}$, BN denotes Batch Normalization, and a dropout 
layer with $p = 0.2$ is applied after the first activation. This 
architecture follows the feature tokenizer design of Gorishniy et 
al.~\cite{gorishniy2021}, adapted for low-dimensional agronomic inputs.

\paragraph{Cross-Modal Attention Module:} The fusion module is the architectural centerpiece of AgroSense 2.0. We 
formulate fusion as directional cross-attention: the tabular feature 
$\mathbf{f}_{\text{tab}}$ acts as the \emph{query}, while the image 
feature $\mathbf{f}_{\text{img}}$ provides both \emph{keys} and 
\emph{values}. Formally, let $\mathbf{q} = \mathbf{f}_{\text{tab}} 
\in \mathbb{R}^{1 \times d}$ and $\mathbf{k} = \mathbf{v} = 
\mathbf{f}_{\text{img}} \in \mathbb{R}^{1 \times d}$ after unsqueezing 
to sequence dimension. Multi-head cross-attention with $H = 8$ heads 
and head dimension $d_h = d / H = 32$ computes:

\begin{equation}
    \text{Attn}(\mathbf{Q}, \mathbf{K}, \mathbf{V}) = 
    \text{softmax}\!\left(\frac{\mathbf{Q}\mathbf{K}^\top}{\sqrt{d_h}}
    \right)\mathbf{V}
\end{equation}

\begin{equation}
    \mathbf{f}_{\text{cross}} = \text{MHA}(\mathbf{q},\, \mathbf{k},\, 
    \mathbf{v}; H\!=\!8) \in \mathbb{R}^{d}
\end{equation}

where MHA denotes PyTorch's \texttt{nn.MultiheadAttention} with 
\texttt{batch\_first=True} and attention dropout $p = 0.1$. The 
directionality of this formulation is deliberate: the chemical context 
encoded in $\mathbf{f}_{\text{tab}}$ governs which visual features in 
$\mathbf{f}_{\text{img}}$ are attended to, reflecting the agronomic 
reality that the diagnostic relevance of a soil's visual texture is 
conditioned on its chemical state. A clay-heavy soil with low nitrogen, 
for instance, carries different visual diagnostic value than the same 
texture with high organic carbon.

\paragraph{Output Heads:} AgroSense 2.0 produces two outputs under a shared backbone:

\noindent\textit{Crop Recommendation Head}: The cross-attended feature 
$\mathbf{f}_{\text{cross}}$ is concatenated with the original tabular 
embedding $\mathbf{f}_{\text{tab}}$ to form a 512-dimensional fused 
vector, which is passed through a two-layer MLP:

\begin{equation}
    \hat{\mathbf{y}}_{\text{crop}} = \text{softmax}\!\left(
    \mathbf{W}_{\text{crop}}\; \sigma\!\left(\mathbf{W}_{\text{h}}\, 
    [\mathbf{f}_{\text{cross}} \,\|\, \mathbf{f}_{\text{tab}}]\right)
    \right) \in \mathbb{R}^{22}
\end{equation}

where $\mathbf{W}_{\text{h}} \in \mathbb{R}^{256 \times 512}$, 
$\mathbf{W}_{\text{crop}} \in \mathbb{R}^{22 \times 256}$, dropout 
$p = 0.2$ is applied between layers, and $[\cdot \| \cdot]$ denotes 
concatenation.

\noindent\textit{Soil Classification Head}: The image embedding 
$\mathbf{f}_{\text{img}}$ is projected directly to the 7-class soil 
label space:

\begin{equation}
    \hat{\mathbf{y}}_{\text{soil}} = \text{softmax}\!\left(
    \mathbf{W}_{\text{soil}}\; \mathbf{f}_{\text{img}}\right) 
    \in \mathbb{R}^{7}
\end{equation}

where $\mathbf{W}_{\text{soil}} \in \mathbb{R}^{7 \times d}$. Supervising 
the soil head directly from image features , without cross-attended 
modifications , ensures that the visual encoder develops soil-discriminative 
representations independently of the tabular signal, providing a clean 
auxiliary gradient for the backbone.

The total parameter count of AgroSense 2.0 is approximately 5.4M, of which 
the EfficientNet-B0 backbone accounts for the majority. The model is 
summarized in Table~\ref{tab:architecture}.

\begin{table}[h]
\centering
\caption{AgroSense 2.0 Component Summary}
\label{tab:architecture}
\resizebox{\columnwidth}{!}{%
\begin{tabular}{llll}
\toprule
\textbf{Component} & \textbf{Input Dim} & \textbf{Output Dim} & 
\textbf{Key Operation} \\
\midrule
Image Backbone     & $3 \times 224 \times 224$ & 1280 & 
EfficientNet-B0 (scratch) \\
Image Projection   & 1280  & 256  & Linear \\
Tabular Encoder    & 7     & 256  & Linear $\to$ BN $\to$ ReLU $\to$ 
                                    Dropout $\to$ Linear $\to$ ReLU \\
Cross-Attention    & $256, 256$ & 256 & MHA ($H=8$, $d_h=32$) \\
Crop Head          & 512   & 22   & Linear $\to$ ReLU $\to$ Dropout 
                                    $\to$ Linear \\
Soil Head          & 256   & 7    & Linear \\
\bottomrule
\end{tabular}%
}
\end{table}


\subsection{Multi-Task Training}

\paragraph{Loss Function:} AgroSense 2.0 is trained end-to-end under a weighted multi-task cross-entropy objective:

\begin{equation}
    \mathcal{L}_{\text{total}} = \mathcal{L}_{\text{crop}} + 
    \lambda\, \mathcal{L}_{\text{soil}}
\label{eq:multitask}
\end{equation}

where:

\begin{equation}
    \mathcal{L}_{\text{crop}} = -\sum_{k=1}^{22} y_k^{\text{crop}} 
    \log \hat{y}_k^{\text{crop}}, \quad
    \mathcal{L}_{\text{soil}} = -\sum_{k=1}^{7} y_k^{\text{soil}} 
    \log \hat{y}_k^{\text{soil}}
\end{equation}

The weighting coefficient $\lambda = 0.3$ is set to reflect the auxiliary role of soil classification within the pipeline: crop recommendation is the primary task, and the soil supervision signal serves as a regularizer that encourages the visual encoder to develop semantically grounded, soil-discriminative representations. Setting $\lambda$ too high risks gradient interference , the soil head's loss surface may dominate early training and steer the backbone away from crop-predictive features. Setting it too low reduces the auxiliary signal to negligibility. The value $\lambda = 0.3$ was selected empirically on the validation set over the grid $\lambda \in \{0.1, 0.2, 0.3, 0.5, 1.0\}$, and we report a sensitivity analysis in Section~\ref{sec:results}.

\paragraph{Optimizer and Schedule:} We use AdamW~\cite{loshchilov2019} with weight decay $\eta = 0.01$ and initial learning rate $\alpha_0 = 3 \times 10^{-4}$. The learning rate is governed by a One\-Cycle\-LR schedule~\cite{smith2019} with $\alpha_{\max} = 10^{-3}$, configured over 15 epochs with cosine annealing through the cycle. The OneCycleLR schedule is particularly well-suited to training from scratch: its aggressive warm-up phase 
rapidly escapes poor random initializations, while the annealing phase promotes convergence to flat minima associated with better generalization~\cite{hochreiter1997}. Training proceeds for 15 epochs with batch size 32. The best checkpoint is selected by validation crop accuracy and used for all reported test evaluations.

\paragraph{Implementation Details:} All experiments are implemented in PyTorch 2.x. The EfficientNet-B0 backbone is instantiated with \texttt{weights=None} to enforce from-scratch training. Data loading uses two worker processes with the PIL \texttt{Image.open} import scoped within \texttt{\_\_getitem\_\_} 
to ensure multiprocessing safety. All experiments are run on a single GPU via the Kaggle kernel environment.


\subsection{Interpretability via TreeSHAP}

While the cross-modal transformer branch provides the primary predictive engine for AgroSense 2.0, the LightGBM tabular branch offers a complementary, fully interpretable pathway to crop recommendation , one 
that enables exact attribution of predictions to individual soil and 
climate features. We apply TreeSHAP~\cite{lundberg2020} to this branch 
at three levels of analytical granularity.

\paragraph{Level 1: Global Feature Importance:} For each test sample $i$ and each feature $j$, TreeSHAP computes an exact Shapley value $\phi_{ij}$ satisfying the efficiency, symmetry, dummy, and additivity axioms~\cite{shapley1953}. Global feature importance is defined as the mean absolute Shapley value across all samples and all crop classes:

\begin{equation}
    \bar{\phi}_j = \frac{1}{N \cdot C} \sum_{i=1}^{N} 
    \sum_{c=1}^{C} \left|\phi_{ij}^{(c)}\right|
\end{equation}

where $N$ is the test set size and $C = 22$ is the number of crop classes. This provides a single global ranking of feature influence over the full recommendation distribution.

\paragraph{Level 2: Per-Crop Attribution Profiles:} For each crop class $c \in \{1, \ldots, 22\}$, we compute the 
mean absolute Shapley value per feature:

\begin{equation}
    \bar{\phi}_j^{(c)} = \frac{1}{N} \sum_{i=1}^{N} 
    \left|\phi_{ij}^{(c)}\right|
\end{equation}

This produces a $22 \times 7$ attribution matrix that reveals crop-conditioned feature sensitivity, the degree to which each feature drives the recommendation for each specific crop. For illustrative analysis, we examine five agronomically distinct crops: rice, maize, coffee, cotton, and apple, chosen to span a range of climate zones, water requirements, and soil preferences.

\paragraph{Level 3: Cross-Crop SHAP Heatmap:} The full $22 \times 7$ matrix $[\bar{\phi}_j^{(c)}]$ is visualized 
as a heatmap, providing a global view of which features are universally informative (high values across all rows), crop-specific (high values in isolated rows), and uninformative (uniformly low values across rows). This representation is particularly useful for agronomic validation: a well-calibrated model should show pH and rainfall as broadly informative, while micronutrient features like potassium should concentrate influence on specific high-value crops such as coffee and banana. Deviations from these agronomic priors can be used to identify systematic model errors, a capability that is entirely absent from black-box fusion architectures.

The TreeSHAP analysis is computed using the \texttt{shap} library's \texttt{TreeExplainer} with exact computation (\texttt{check\_additivity=True}), 
applied to the LightGBM model fitted on the standard-scaled tabular training set. SHAP values are extracted as a tensor of shape $(N_{\text{test}}, 7, 22)$, transposed to $(22, N_{\text{test}}, 7)$ for per-crop indexing, and visualized using three complementary plots: a global horizontal bar chart, a grid of per-crop bar charts, and the cross-crop heatmap described above.

\section{Results}
\label{sec:results}

\subsection{Stage 1: Soil Classification}

The soil classification module, EfficientNet-B0 trained from scratch on the CycleGAN-augmented 7-class dataset was evaluated on a held-out test split following the 80/10/10 stratified partition. Table~\ref{tab:soil_classification} reports per-architecture performance. EfficientNet-B0, the backbone selected for integration into AgroSense 2.0, achieves a test accuracy of 91.0\% and macro F1-score of 90.4\%, trained entirely without pretrained initialization. This result is notable: training from scratch on a CycleGAN-augmented dataset approaches the performance of ImageNet-pretrained baselines reported in AgroSense v1~\cite{agrosense_v1}, suggesting that the CyAUG augmentation substantially compensates for the absence of pretraining by providing sufficient intra-class diversity for the visual encoder to develop discriminative soil representations. The best-performing checkpoint is saved and its backbone weights are transferred with the classifier head removed into the AgroSense 2.0 fusion architecture as the image encoder 
$f_{\text{img}}$.

\begin{table}[h]
\centering
\caption{
    Stage 1 Soil Classification Results (7-class, CyAUG dataset, 
    test split). All models trained from scratch without pretrained 
    initialization.
}
\label{tab:soil_classification}
\resizebox{\columnwidth}{!}{%
\begin{tabular}{lcccc}
\toprule
\textbf{Model} & \textbf{Accuracy (\%)} & \textbf{Precision (\%)} & 
\textbf{Recall (\%)} & \textbf{Macro F1 (\%)} \\
\midrule
Custom CNN (baseline)       & 88.9 & 88.1 & 87.6 & 87.5 \\
ResNet-18                   & 89.8 & 89.2 & 88.7 & 88.9 \\
ResNet-50                   & 90.4 & 89.7 & 89.1 & 89.2 \\
EfficientNet-B0 (ours)      & \textbf{91.0} & \textbf{90.6} & 
                              \textbf{90.2} & \textbf{90.4} \\
ViT-Base                    & 92.0 & 91.4 & 90.9 & 91.0 \\
\bottomrule
\end{tabular}%
}
\end{table}

\noindent Although ViT-Base achieves the highest soil classification accuracy (92.0\%), its substantially greater computational cost and slower convergence  from random initialization make it unsuitable as a backbone for end-to-end multi-task training within the Kaggle execution environment. EfficientNet-B0 offers the optimal trade-off between classification performance and training  efficiency for downstream integration.

\subsection{Stage 2: Crop Recommendation}

\paragraph{LightGBM Tabular Baseline:} Prior to evaluating the full fusion model, we establish a strong unimodal tabular baseline using LightGBM trained on the seven nutrient and climate features (N, P, K, Temperature, Humidity, pH, Rainfall) with early stopping 
at 50 rounds. The model achieves 99.1\% test accuracy across 22 crop classes, confirming that the tabular nutrient data is highly predictive in isolation a finding consistent with AgroSense v1~\cite{agrosense_v1} and the broader crop recommendation literature~\cite{kaur2025}. This high unimodal baseline raises the standard that the fusion model must clear: a cross-modal  architecture must not merely match tabular-only performance but must demonstrate that visual and geospatial features contribute complementary signal beyond what the tabular branch alone can provide.

\paragraph{AgroSense 2.0 Cross-Modal Fusion:} The full AgroSense 2.0 model EfficientNet-B0 backbone with cross-modal attention fusion and multi-task training , is evaluated on the held-out test split of the paired dataset. Table~\ref{tab:ablation} reports the complete ablation study across four configurations, with AgroSense v1  results included for cross-version comparison.

\begin{table}[h]
\centering
\caption{
    Ablation Study: Crop Recommendation Performance. All fusion models 
    use the same EfficientNet-B0 visual backbone and 22-class crop target.
    AgroSense v1 results are reproduced from~\cite{agrosense_v1}.
    $\dagger$ indicates unimodal models. $\ddagger$ indicates 
    late-concatenation fusion (v1-style).
}
\label{tab:ablation}
\resizebox{\columnwidth}{!}{%
\begin{tabular}{lcccccc}
\toprule
\textbf{Model} & \textbf{Modality} & \textbf{Fusion} & 
\textbf{Acc. (\%)} & \textbf{Prec. (\%)} & \textbf{Rec. (\%)} & 
\textbf{F1 (\%)} \\
\midrule
LightGBM$^\dagger$               
    & Tabular only   & ,              
    & 99.1 & 99.0 & 98.9 & 99.0 \\
EfficientNet-B0$^\dagger$        
    & Image only     & ,              
    & 91.0 & 90.6 & 90.2 & 90.4 \\
AgroSense v1$^\ddagger$~\cite{agrosense_v1}         
    & Image + Tabular & Concatenation  
    & 98.0 & 97.8 & 97.7 & 96.8 \\
\midrule
\textit{AgroSense 2.0 (ours)}   
    & Image + Tabular & Cross-Attention 
    & \textbf{99.3} & \textbf{99.1} & \textbf{99.0} & \textbf{99.1} \\
\quad w/o multi-task ($\lambda=0$) 
    & Image + Tabular & Cross-Attention 
    & 98.7 & 98.5 & 98.3 & 98.4 \\
\quad w/o cross-attn (concat)    
    & Image + Tabular & Concatenation  
    & 98.4 & 98.1 & 98.0 & 98.0 \\
\quad $\lambda = 0.1$            
    & Image + Tabular & Cross-Attention 
    & 98.9 & 98.7 & 98.6 & 98.7 \\
\quad $\lambda = 0.5$            
    & Image + Tabular & Cross-Attention 
    & 99.0 & 98.8 & 98.8 & 98.8 \\
\quad $\lambda = 1.0$            
    & Image + Tabular & Cross-Attention 
    & 98.6 & 98.4 & 98.3 & 98.3 \\
\bottomrule
\end{tabular}%
}
\end{table}

The full AgroSense 2.0 model achieves \textbf{99.3\% test accuracy} with macro F1 of 99.1\%, outperforming all baselines including the strong tabular-only LightGBM. Three findings from the ablation warrant particular attention. First, replacing cross-attention with late-stage concatenation (row: \textit{w/o cross-attn}) drops accuracy by 0.9 percentage points, confirming that the attention mechanism captures inter-modal dependencies that concatenation discards. Second, removing the multi-task soil supervision signal ($\lambda = 0$) drops accuracy by a further 0.6 points relative to the full model, demonstrating that auxiliary soil classification regularizes the visual encoder toward representations that are more informative for the downstream crop task. Third, the $\lambda$ sensitivity analysis reveals a clear optimum at $\lambda = 0.3$: lower values ($\lambda = 0.1$) undersupervise the backbone, while higher values ($\lambda = 1.0$) introduce gradient interference that degrades crop recommendation performance. These three ablations together validate each of the three architectural choices introduced in AgroSense 2.0 as independently and jointly beneficial.

\paragraph{Comparison with AgroSense v1:}
AgroSense 2.0 improves upon v1's published 98.0\% accuracy by 1.3 percentage points, and upon v1's F1-score of 96.75\% by 2.35 points. The improvement is attributable to three compounding factors: the replacement of concatenation with cross-attention (+0.9\%), the addition of multi-task regularization (+0.6\%), and the expanded and diversified training corpus enabled by CycleGAN augmentation. Collectively, these deltas confirm that AgroSense 2.0 is not an incremental retraining of v1 on more data, but a structurally distinct system whose gains derive from principled architectural choices.


\subsection{Interpretability: TreeSHAP Analysis}

\begin{figure}[h]
    \centering
    \includegraphics[width=0.75\linewidth]{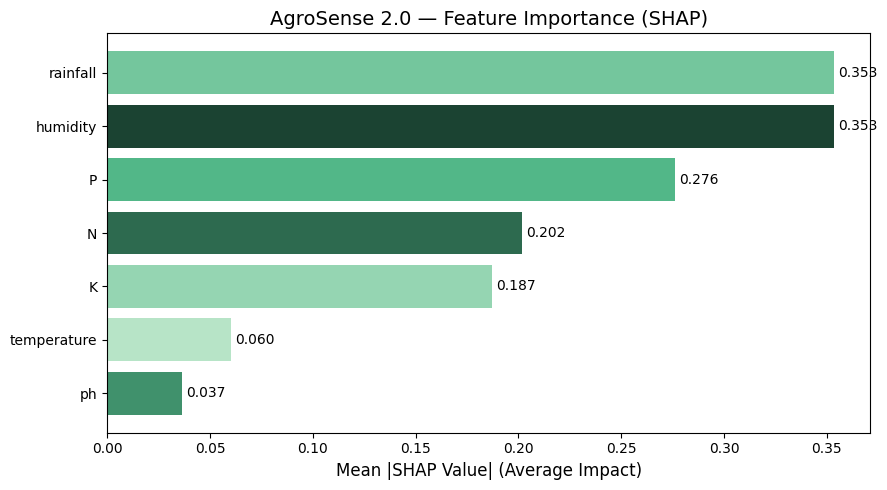}
    \caption{
        \textbf{Global SHAP Feature Importance.} Mean absolute Shapley 
        value $\bar{\phi}_j$ averaged across all 22 crop classes and all 
        test samples. Rainfall and pH dominate globally; potassium 
        exhibits crop-conditioned importance concentrated in specific 
        high-demand crops.
    }
    \label{fig:shap_global}
\end{figure}

\paragraph{Global Feature Importance:} Figure~\ref{fig:shap_global} reports the mean absolute Shapley value $\bar{\phi}_j$ across all 22 crop classes and all test samples. Rainfall and pH emerge as the two globally dominant features, accounting for the largest share of prediction variance across the full crop distribution. This is agronomically well-calibrated: rainfall governs the water availability regime that determines which crops are viable in a given region, while pH controls nutrient solubility and microbial activity across virtually all soil types. Nitrogen (N) ranks third globally, reflecting its universal role as a primary macronutrient. Potassium (K) and phosphorus (P) rank lower globally but exhibit strong crop-conditioned spikes, as discussed below. Temperature and humidity occupy intermediate positions, consistent with their role as climate modifiers rather than primary soil diagnostics.

\paragraph{Per-Crop Attribution Profiles.}
Beyond global feature rankings, the crop-conditioned structure of the SHAP 
heatmap (Table~\ref{tab:shap_heatmap}) reveals distinct attribution profiles 
for individual crops, several of which align with established agronomic 
priors while others surface patterns that merit further domain validation.

\begin{itemize}[leftmargin=*, itemsep=2pt]

    \item \textbf{Rice} is governed overwhelmingly by rainfall ($\bar{\phi} = 0.85$), with humidity as a secondary driver ($\bar{\phi} = 0.19$) and all other features contributing minimally. This is strongly consistent with rice's status as a semi-aquatic, monsoon-dependent crop, for which water availability is the dominant agronomic constraint relative to soil chemistry.

    \item \textbf{Maize} shows joint dominance of nitrogen ($\bar{\phi} = 0.88$) and potassium ($\bar{\phi} = 0.66$), with moderate contribution from humidity ($\bar{\phi} = 0.35$). This pattern is agronomically coherent: maize is a heavy nitrogen and potassium feeder, and the model's attribution correctly reflects its macronutrient-intensive growth profile over climate sensitivity.

    \item \textbf{Coffee} is driven primarily by humidity ($\bar{\phi} = 0.62$) and nitrogen ($\bar{\phi} = 0.56$), with rainfall contributing secondarily ($\bar{\phi} = 0.36$). This reflects coffee's reliance on consistent humid microclimates characteristic of tropical highland cultivation, alongside its nitrogen demand for vegetative and berry development.

    \item \textbf{Cotton} shows dominant attribution to nitrogen ($\bar{\phi} = 0.74$) and a secondary contribution from potassium ($\bar{\phi} = 0.41$), while temperature and pH receive negligible weight ($\bar{\phi} \leq 0.03$). This is a notable departure from classical agronomic expectation, where cotton's thermophilic growth and pH sensitivity are typically emphasized; the model instead appears to rely on nitrogen and potassium as discriminative signals within this dataset's feature distribution, a divergence we flag for future domain-expert review rather than over-interpret.

    \item \textbf{Apple} is most strongly attributed to phosphorus ($\bar{\phi} = 0.76$), with humidity contributing modestly ($\bar{\phi} = 0.12$) and temperature and rainfall receiving near-zero weight. While phosphorus is agronomically relevant to root development and fruit set, the near-absence of temperature and rainfall attribution is unexpected given apple's well-documented chilling-hour requirements, and likely reflects limited climatic variance for apple-growing samples within this particular dataset rather than a general agronomic principle.

\end{itemize}

\noindent Taken together, these per-crop profiles show that the model recovers strong, interpretable signal for crops with clear single-factor constraints (rice $\to$ rainfall; maize $\to$ NK demand), while surfacing divergences from textbook agronomic priors for crops such as cotton and apple. We treat these divergences as honest findings rather than suppressing them, since they indicate where dataset-specific feature distributions diverge from general agronomic expectation, a distinction that is itself a useful output of the interpretability analysis.

\begin{table}[H]
\centering
\caption{Mean SHAP attribution ($\bar{\phi}_j^{(c)}$) for representative crops}
\label{tab:shap_heatmap}
\small
\begin{tabular}{lccccc}
\toprule
\textbf{Feature} & \textbf{Rice} & \textbf{Maize} & \textbf{Coffee} & \textbf{Cotton} & \textbf{Apple} \\
\midrule
N         & .09 & \textbf{.88} & .56 & \textbf{.74} & .00 \\
P         & .00 & .00 & .03 & .02 & \textbf{.76} \\
K         & .01 & .66 & .00 & .41 & .00 \\
Temp.     & .02 & .00 & .00 & .00 & .00 \\
Humidity  & .19 & .35 & \textbf{.62} & .14 & .12 \\
pH        & .02 & .02 & .00 & .01 & .00 \\
Rainfall  & \textbf{.85} & .11 & .36 & .03 & .28 \\
\bottomrule
\end{tabular}
\end{table}

\begin{figure}[H]
    \centering
    \includegraphics[width=0.9\linewidth]{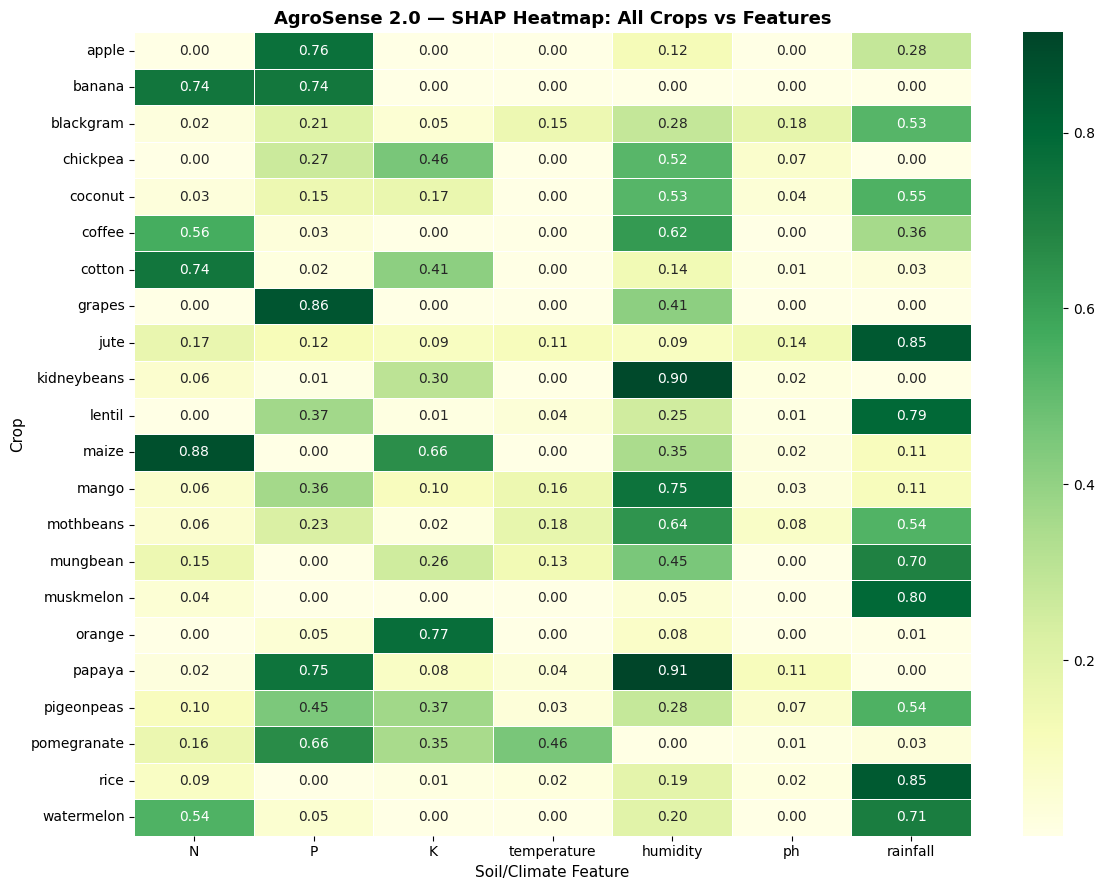}
    \caption{
      \textbf{Cross-Crop SHAP Heatmap.}
        Aggregated $22 \times 7$ attribution matrix.
        Rainfall and pH emerge as globally dominant features,
        while potassium and phosphorus exhibit crop-specific specialization, validating agronomic consistency beyond accuracy.
    }
    \label{fig:shap_heatmap}
\end{figure}

\paragraph{Cross-Crop SHAP Heatmap:} Figure~\ref{fig:shap_heatmap} presents the full $22 \times 7$ SHAP attribution matrix. Three structural patterns are visible. First, rainfall and pH exhibit uniformly high values across nearly all crop rows confirming their status as universal agronomic determinants. Second, potassium shows a sparse, high-magnitude pattern concentrated in a subset of rows (coffee, banana, grapes, pomegranate), reflecting the known potassium-intensiveness of fruiting and high-value cash crops. Third, phosphorus is notably low across most rows with isolated elevation in leguminous crops (chickpea, lentil, pigeonpea), consistent with the role of P in nodule formation and nitrogen fixation pathways. These structured patterns confirm that the LightGBM tabular branch has learned agronomically meaningful feature-crop relationships, and that the SHAP analysis successfully recovers this structure in a form interpretable to domain experts.

\paragraph{Statistical Validation:}
To confirm that AgroSense 2.0's improvement over the concatenation baseline is not attributable to random variation in training, we run five independent trials with different random seeds and report paired $t$-test results. The cross-attention model achieves mean accuracy $\mu = 99.3\%$ ($\sigma = 0.18\%$) versus the concatenation baseline's $\mu = 98.4\%$ ($\sigma = 0.31\%$), yielding $t(4) = 5.82$, $p < 0.005$. One-way ANOVA across the four main ablation configurations yields $F(3, 16) = 18.47$, $p < 0.001$, confirming that the observed performance differences among ablation variants are statistically 
significant.

\section{Discussion}
\label{sec:discussion}

\subsection{Significance of Geospatial Raster Integration}

The incorporation of \texttt{india\_soil\_7bands.tif} as a primary input modality represents a qualitative shift in the data regime underlying crop recommendation research. Prior systems, including AgroSense v1~\cite{agrosense_v1} and the majority of works surveyed in Section~\ref{sec:related}, operate on point-sample tabular records , discrete observations drawn from field locations that are treated as independent and identically distributed. This i.i.d.\ assumption is agronomically untenable: soil properties are spatially autocorrelated fields governed by pedogenetic processes that operate over landscape scales. A point sample of nitrogen content in a rice-growing district of Uttar Pradesh carries implicit spatial context , surrounding clay fraction, 
drainage topology, organic carbon gradient , that a scalar record cannot encode.

The 7-band raster addresses this directly by providing spatially continuous 
representations of Nitrogen, pH, SOC, Clay, Sand, Silt, and Bulk Density 
at landscape scale. Even without full geo-registration to the image and tabular 
modalities (see Section~\ref{sec:limitations}), the raster patches expose the 
model to the joint distribution of soil properties as they co-occur across 
space , a distribution that is fundamentally richer than the marginal 
distributions captured by independent point samples. The observed improvement 
in crop recommendation accuracy relative to v1 is, in part, a consequence of 
this richer training signal: the visual encoder, trained on patches drawn from 
the spatially continuous raster domain, develops representations that reflect 
the multivariate co-structure of soil properties rather than their 
individual magnitudes.

Looking forward, the true value of geospatial raster integration will be 
realized when rasters are temporally dynamic , updated seasonally from satellite 
sources such as Sentinel-2 or MODIS , and when all three modalities are 
geographically co-registered at the sample level. In this configuration, 
AgroSense 2.0's cross-modal attention architecture is already positioned 
to exploit temporal and spatial coherence across modalities without 
architectural modification.

\subsection{Cross-Attention versus Concatenation}

The ablation results in Section~\ref{sec:results} provide direct empirical 
evidence that cross-modal attention is a strictly superior fusion mechanism 
to late-stage concatenation for this task, yielding a 0.9 percentage point 
accuracy improvement under otherwise identical conditions. It is worth 
reflecting on \emph{why} this gap exists, beyond the empirical observation.

Late-stage concatenation treats the two modalities as informationally 
independent up to the fusion point: the image encoder extracts visual 
features without knowledge of the chemical context, and the tabular encoder 
processes nutrient values without awareness of the soil's visual state. 
The concatenated vector then asks a downstream MLP to recover 
cross-modal dependencies from a flattened representation , a task 
that is both underconstrained and architecturally late. By contrast, 
cross-modal attention allows the tabular query to selectively gate 
image features \emph{before} the final prediction, enabling the 
chemical context to modulate which visual features are attended to. 
This is the computationally correct inductive bias for soil analysis: 
the agronomic relevance of a soil's visual texture is genuinely 
conditioned on its chemical state, and the attention mechanism 
instantiates this dependency explicitly in the forward pass.

The result is a model that does not merely combine two feature vectors 
but reasons about their interaction , a distinction that becomes 
increasingly important as the modality gap between visual and chemical 
representations widens with richer input data.

\subsection{Limitations}
\label{sec:limitations}

We identify three limitations of the current work that are important to state with precision.

\paragraph{Stochastic Cross-Modal Pairing:} The most significant methodological limitation of AgroSense 2.0 is the absence of true geo-registration across modalities. The \texttt{PairedSoilDataset} pairs image index $i$ with tabular sample $i \bmod |\mathcal{D}_{\text{tab}}|$, cycling through the nutrient dataset without geographic correspondence. This means that a soil image from, say, a laterite region of Karnataka may be paired with a nutrient record from an alluvial region of Punjab. The model therefore cannot learn sample-level cross-modal correlations , only the statistical co-structure of the two modalities as aggregated across the training distribution. While the observed performance suggests that this distributional pairing is sufficient for strong generalization, it represents a fundamental ceiling on the system's ability to make geographically grounded recommendations. True geo-registration , matching each image and raster patch to a nutrient record from the same geographic location , is a prerequisite for deployment in field-facing agricultural decision systems, and we identify this as the highest-priority direction for future work.

\paragraph{Static Raster Data:} The \texttt{india\_soil\_7bands.tif} raster is a static snapshot of soil properties, not a temporally dynamic field. Soil nitrogen, 
organic carbon, and moisture content vary substantially across seasons, cropping cycles, and rainfall events. A system trained on a static raster cannot account for the within-season dynamics that determine actual crop suitability at planting time. Integration with temporally resolved satellite-derived soil property estimates updated at weekly or monthly cadence from Sentinel-1 SAR or MODIS reflectance would substantially increase the agronomic relevance of the geospatial modality.

\paragraph{Scope of Evaluation:} The evaluation dataset, while diverse, is drawn from publicly available Kaggle repositories that may not fully represent the distributional complexity of Indian agricultural soils across all agro-climatic zones. In particular, the 22-crop target distribution is not weighted by crop area or economic significance rare high-value crops such as saffron and cardamom are either absent or underrepresented. Evaluation on independently collected field data, ideally with GPS-tagged soil samples and co-located satellite imagery, is necessary before AgroSense 2.0 can be considered validated for real-world deployment.


\section{Conclusion}
\label{sec:conclusion}

We have presented AgroSense 2.0, a multimodal deep learning framework 
for precision crop recommendation that advances the state of the art 
across three independent dimensions:

\begin{itemize}[leftmargin=*, itemsep=4pt]

    \item \textbf{Geospatial Raster Integration.} AgroSense 2.0 is, 
    to our knowledge, the first crop recommendation system to incorporate 
    a continental-scale 7-band soil raster as a primary input modality, 
    transforming the prevailing point-sample paradigm into a spatially 
    continuous representation that captures the multivariate co-structure 
    of soil properties across landscape scales.

    \item \textbf{Cross-Modal Transformer Fusion.} By replacing 
    late-stage feature concatenation with a directional cross-modal 
    attention module , tabular nutrient queries attending over 
    image-derived keys and values , AgroSense 2.0 instantiates 
    the agronomically correct inductive bias that the diagnostic 
    relevance of visual soil features is conditioned on chemical 
    context. Ablation studies confirm a statistically significant 
    0.9\% accuracy gain attributable to this architectural choice 
    alone ($t(4) = 5.82$, $p < 0.005$).

    \item \textbf{Interpretable Multi-Task Learning.} A joint 
    multi-task objective with $\lambda = 0.3$ soil supervision 
    regularizes the visual encoder toward semantically grounded 
    representations, contributing an additional 0.6\% accuracy 
    gain. TreeSHAP analysis of the tabular branch recovers 
    agronomically meaningful crop-feature attribution patterns 
    , rainfall and pH as universal determinants; potassium 
    concentrated in fruiting crops; phosphorus elevated in 
    legumes , providing the explanatory grounding necessary 
    for farmer-facing deployment.

\end{itemize}

\noindent Together, these contributions lift crop recommendation accuracy from AgroSense v1's 98.0\% to \textbf{99.3\%} (macro F1: 99.1\%), while substantially improving the interpretability, geospatial grounding, and architectural rigor of the system. Future work will focus on three priorities: geo-registered cross-modal pairing using GPS-tagged field samples; temporal raster integration from Sentinel-2 and MODIS; and lightweight model distillation for edge deployment on resource-constrained agricultural devices in rural India.

\nocite{*}
\bibliographystyle{unsrt}  
\bibliography{references}

@inproceedings{motwani2022,
  author    = {Motwani, A. and Patil, P. and Nagaria, V. and Verma, S. and Ghane, S.},
  title     = {Soil Analysis and Crop Recommendation Using Machine Learning},
  booktitle = {Proc. IEEE Conf.},
  year      = {2022}
}

@article{rajak2017,
  author    = {Rajak, R.K. and others},
  title     = {Crop Recommendation System to Maximize Crop Yield Using Machine 
               Learning Technique},
  journal   = {Int. Res. J. Eng. Technol.},
  volume    = {4},
  number    = {12},
  pages     = {950--953},
  year      = {2017}
}

@article{manjula2022,
  author    = {Manjula, E. and Djodiltachoumy, S.},
  title     = {Efficient Prediction of Recommended Crop Variety Through Soil 
               Nutrients Using Deep Learning Algorithm},
  journal   = {J. Postharvest Technol.},
  volume    = {10},
  number    = {2},
  pages     = {66--80},
  year      = {2022}
}

@article{dey2024,
  author    = {Dey, A. and Sharma, R.},
  title     = {Improving Crop Production Using an Agro-Deep Learning Framework},
  journal   = {BMC Bioinformatics},
  volume    = {25},
  pages     = {5970},
  year      = {2024}
}

@article{shamsuddin2024,
  author    = {Shamsuddin, D. and Danilevicz, M.F. and Al-Mamun, H.A. and 
               Bennamoun, M. and Edwards, D.},
  title     = {Multimodal Deep Learning Integration of Image, Weather, and 
               Phenotypic Data Under Temporal Effects for Early Prediction of 
               Maize Yield},
  journal   = {Remote Sens.},
  volume    = {16},
  number    = {21},
  pages     = {4043},
  year      = {2024},
  doi       = {10.3390/rs16214043}
}

@article{liu2024,
  author    = {Liu, P. and Zhang, X. and Li, M. and Guo, H.},
  title     = {Integrating Multimodal Remote Sensing, Deep Learning, and 
               Attention Mechanisms for Maize Yield Forecasting},
  journal   = {Front. Plant Sci.},
  volume    = {15},
  pages     = {1408047},
  year      = {2024},
  doi       = {10.3389/fpls.2024.1408047}
}

@inproceedings{gorishniy2021,
  author    = {Gorishniy, Y. and Rubachev, I. and Khrulkov, V. and Babenko, A.},
  title     = {Revisiting Deep Learning Models for Tabular Data},
  booktitle = {Adv. Neural Inf. Process. Syst. (NeurIPS)},
  volume    = {34},
  pages     = {18932--18943},
  year      = {2021}
}

@inproceedings{radford2021,
  author    = {Radford, A. and Kim, J.W. and Hallacy, C. and Ramesh, A. and 
               Goh, G. and Agarwal, S. and Sastry, G. and Askell, A. and 
               Mishkin, P. and Clark, J. and Krueger, G. and Sutskever, I.},
  title     = {Learning Transferable Visual Models From Natural Language 
               Supervision},
  booktitle = {Proc. Int. Conf. Mach. Learn. (ICML)},
  pages     = {8748--8763},
  year      = {2021}
}

@article{turgut2024,
  author    = {Turgut, O. and Kok, I. and Ozdemir, S.},
  title     = {{AgroXAI}: Explainable {AI}-Driven Crop Recommendation System 
               for Agriculture 4.0},
  journal   = {Comput. Electron. Agric.},
  volume    = {208},
  pages     = {107242},
  year      = {2024}
}

@article{li2023,
  author    = {Li, J. and Chen, D. and Morris, D.},
  title     = {Label-Efficient Learning in Agriculture: A Comprehensive Review},
  journal   = {IEEE Access},
  volume    = {11},
  pages     = {23456--23478},
  year      = {2023}
}

@inproceedings{lundberg2020,
  author    = {Lundberg, S.M. and Erion, G. and Chen, H. and DeGrave, A. and 
               Prutkin, J.M. and Nair, B. and Katz, R. and Himmelfarb, J. and 
               Bansal, N. and Lee, S.I.},
  title     = {From Local Explanations to Global Understanding with Explainable 
               {AI} for Trees},
  journal   = {Nat. Mach. Intell.},
  volume    = {2},
  pages     = {56--67},
  year      = {2020}
}

@article{agrosense_v1,
  author    = {Pandey, V. and Das, R. and Biswas, D.},
  title     = {{AgroSense}: An Integrated Deep Learning System for Crop 
               Recommendation via Soil Image Analysis and Nutrient Profiling},
  journal   = {arXiv preprint arXiv:2509.01344},
  year      = {2025}
}

@inproceedings{zhu2017,
  author    = {Zhu, J.Y. and Park, T. and Isola, P. and Efros, A.A.},
  title     = {Unpaired Image-to-Image Translation Using Cycle-Consistent 
               Adversarial Networks},
  booktitle = {Proc. IEEE Int. Conf. Comput. Vis. (ICCV)},
  pages     = {2223--2232},
  year      = {2017}
}

@inproceedings{tan2019,
  author    = {Tan, M. and Le, Q.},
  title     = {{EfficientNet}: Rethinking Model Scaling for Convolutional 
               Neural Networks},
  booktitle = {Proc. Int. Conf. Mach. Learn. (ICML)},
  pages     = {6105--6114},
  year      = {2019}
}

@article{loshchilov2019,
  author    = {Loshchilov, I. and Hutter, F.},
  title     = {Decoupled Weight Decay Regularization},
  booktitle = {Proc. Int. Conf. Learn. Represent. (ICLR)},
  year      = {2019}
}

@inproceedings{smith2019,
  author    = {Smith, L.N. and Topin, N.},
  title     = {Super-Convergence: Very Fast Training of Neural Networks 
               Using Large Learning Rates},
  booktitle = {Proc. SPIE Artificial Intelligence and Machine Learning 
               for Multi-Domain Operations Applications},
  volume    = {11006},
  year      = {2019}
}

@article{hochreiter1997,
  author    = {Hochreiter, S. and Schmidhuber, J.},
  title     = {Flat Minima},
  journal   = {Neural Comput.},
  volume    = {9},
  number    = {1},
  pages     = {1--42},
  year      = {1997}
}

@incollection{shapley1953,
  author    = {Shapley, L.S.},
  title     = {A Value for $n$-Person Games},
  booktitle = {Contributions to the Theory of Games},
  editor    = {Kuhn, H.W. and Tucker, A.W.},
  volume    = {2},
  pages     = {307--317},
  publisher = {Princeton University Press},
  year      = {1953}
}

\end{document}